\documentclass{bmvc2k}

\usepackage{graphicx}
\graphicspath{{figures//}}
\usepackage[utf8]{inputenc} 
\usepackage[T1]{fontenc}    
\usepackage{xcolor}
\usepackage{hyperref}       
\usepackage{url}            
\usepackage{booktabs}       
\usepackage{amsfonts}       
\usepackage{nicefrac}       
\usepackage{microtype}      
\usepackage{capt-of}
\usepackage[font=small,labelfont=bf]{caption}
\usepackage[outercaption]{sidecap}
\usepackage{color}
\usepackage{xcolor}
\usepackage{times}
\usepackage{epsfig}
\usepackage{wrapfig}
\usepackage{amsmath}
\usepackage{amssymb}
\usepackage{gensymb}
\usepackage[normalem]{ulem}
\usepackage{placeins}
\usepackage{array,multirow}

\title{Crowd Counting by Adaptively Fusing Predictions from an Image Pyramid}

\addauthor{Di Kang}{dkang5-c@my.cityu.edu.hk}{1}
\addauthor{Antoni Chan}{abchan@cityu.edu.hk}{1}

\addinstitution{
 Department of Computer Science\\
 City University of Hong Kong\\
 Hong Kong SAR
}

\runninghead{Kang, Chan}{Crowd Counting}

\newcommand{\reffig}[1]{Fig.~\ref{#1}}

\newcommand{\mysubsubsection}[1]{{\bf #1:}}

\makeatletter
\g@addto@macro\normalsize{
\setlength\abovedisplayskip{5pt}
\setlength\belowdisplayskip{5pt}
\setlength\abovedisplayshortskip{5pt}
\setlength\belowdisplayshortskip{5pt}
}
\makeatother

\begin{document}

\maketitle

\begin{abstract}
Because of the powerful learning capability of deep neural networks, counting performance via density map estimation has improved significantly during the past several years.
However, it is still very challenging due to severe occlusion, large scale variations, and perspective distortion.
Scale variations (from image to image) coupled with perspective distortion (within one image) result in huge scale changes of the object size.
Earlier methods based on convolutional neural networks (CNN) typically did not handle this scale variation explicitly, until Hydra-CNN and MCNN.
MCNN uses three columns, each with different filter sizes, to extract features at different scales.
In this paper, in contrast to using filters of different sizes, we utilize an image pyramid to deal with scale variations.
It is more effective and efficient to resize the input fed into the network, as compared to using larger filter sizes.
Secondly, we adaptively fuse the predictions from different scales (using adaptively changing per-pixel weights), which makes our method adapt to scale changes within an image.
The adaptive fusing is achieved by generating an across-scale attention map, which softly selects a suitable scale for each pixel, followed by a 1x1 convolution.
Extensive experiments on three popular datasets show very compelling results.

\end{abstract}

\vspace{-0.3cm}
\section{Introduction} \label{sec:intro}
\vspace{-0.2cm}

\par
Automatic analysis of crowded scenes from images has important applications in crowd management, traffic control, urban planning, and surveillance, especially with the rapidly increasing population in major cities.
Crowd counting methods can also be applied to other fields, e.g., cell counting~\cite{Lempitsky2010MESA}, vehicle counting~\cite{Onoro2016Hydra,Kang2017Beyond}, animal migration surveillance~\cite{Arteta2016}.

\par
The number and the spatial arrangement of the crowds are two types of useful information for understanding crowded scenes.
Methods that can simultaneously count and predict the spatial arrangement of the individuals are preferred, since situations where many people are crowded into a small area are very different from those where the same number of people are evenly spread out.
Explicitly detecting every person in the image naturally solves the counting problem and estimates the crowd's spatial information as well. But detection performance decreases quickly as occlusions between objects become severe and as object size becomes smaller. In order to bypass the hard detection problem, regression-based methods~\cite{Chan2008Privacy,Ryan2009,Idrees2013} were developed for the crowd counting task, but did not preserve the spatial information, limiting their usefulness for further crowd analysis, such as detection and tracking.

\par
{\em Object density maps}, originally proposed in~\cite{Lempitsky2010MESA}, preserve both the count and spatial arrangement of the crowd, and have been shown effective at object counting \cite{Lempitsky2010MESA,Zhang2015Cross,Zhang2016MCNN,Onoro2016Hydra,Sam2017Switch,Kang2017ACNN,Sindagi2017CPCNN}, people detection  \cite{Ma2015Small,Kang2017Beyond} and people tracking \cite{Rodriguez2011,Kang2017Beyond,Ren2018Fusing} in crowded scenes.
In an object density map, the integral over any sub-region is the number of objects within that corresponding region in the image.
Density-based methods are better at handling cases where objects are severely occluded, by bypassing the hard detection of every object, while also maintaining some spatial information about the crowd.

\par
Crowd counting is very challenging due to illumination change, severe occlusion, various background, perspective distortion and scale variation, which requires robust and powerful features to overcome these difficulties.
The current state-of-the-art methods for object counting use deep learning to estimate crowd density maps~\cite{Zhang2015Cross,Zhang2016MCNN,Onoro2016Hydra,Walach2016Boost,Sam2017Switch,Kang2017Beyond,Kang2017ACNN,Sindagi2017CPCNN}.
In contrast to image classification, where the main objects are roughly centered, cover a large portion of the image, and exhibit no severe scale variations, in crowd counting, the people appear at very different scales in the same image due to the orientation of the camera.
The detection and image segmentation tasks often require some special designs intending to better handle scale variation, such as image pyramids \cite{Hu2017Tiny,Onoro2016Hydra,Chen2016Attention}, multi-scale feature fusion \cite{Pinheiro2016SharpMask,Lin2017FPN,Zhang2016MCNN,Sam2017Switch} or special network layers, such as spatial transformer layer~\cite{Jaderberg2015Spatial} and deformable convolution layer~\cite{dai17dcn}.
For the counting task, several methods for handling large scale variations have been proposed.
Patch size normalization~\cite{Zhang2015Cross} resizes each image patch so that the objects inside become roughly a fixed scale, but requires the image perspective map.
With Hydra-CNN~\cite{Onoro2016Hydra}, the image patch along with its smaller center crops are all resized to the same fixed size, and fed into different columns of the Hydra-CNN.
In \cite{Zhang2016MCNN,Sam2017Switch,Sindagi2017CPCNN}, multi-column CNN (MCNN) uses different filter sizes for each column  to extract multi-scale features.
ACNN~\cite{Kang2017ACNN} handles scale variation by dynamically generating convolution filter weights according to side information, such as perspective value, camera angle and height.

\par
In this paper, we consider an alternative approach to handling multiple scales based on image pyramids. We first design a backbone fully convolutional network (FCN) with a reasonable receptive field size and depth, serving as a strong baseline.
Secondly, to handle scale variations within the same image, we construct an image pyramid of the input image and run each image through the FCN to obtain predicted density maps at different scales.
Finally, the density map predictions at different scales are fused adaptively at every pixel location, which allows switching predictions between scales within the same image.
The adaptive fusion is obtained by using a sub-network to predict an across-scale attention map~\cite{Chen2016Attention}, followed by a 1$\times$1 convolution.
Our method is tested on three popular crowd counting datasets and shows very compelling performance, while having faster than real-time performance.

\vspace{-0.3cm}
\section{Related works}
\vspace{-0.2cm}

\begin{figure}[tb]
\centering
\small
\begin{tabular}{@{}c@{\hspace{.5mm}}c@{\hspace{.5mm}}c@{}}
  (a) image (1203.7) & (b) density (S1) & (c) density (S2)  \\
  \includegraphics[width=0.27\textwidth]{ShanghaiTech_A_image_121.jpg} &
  \includegraphics[width=0.27\textwidth]{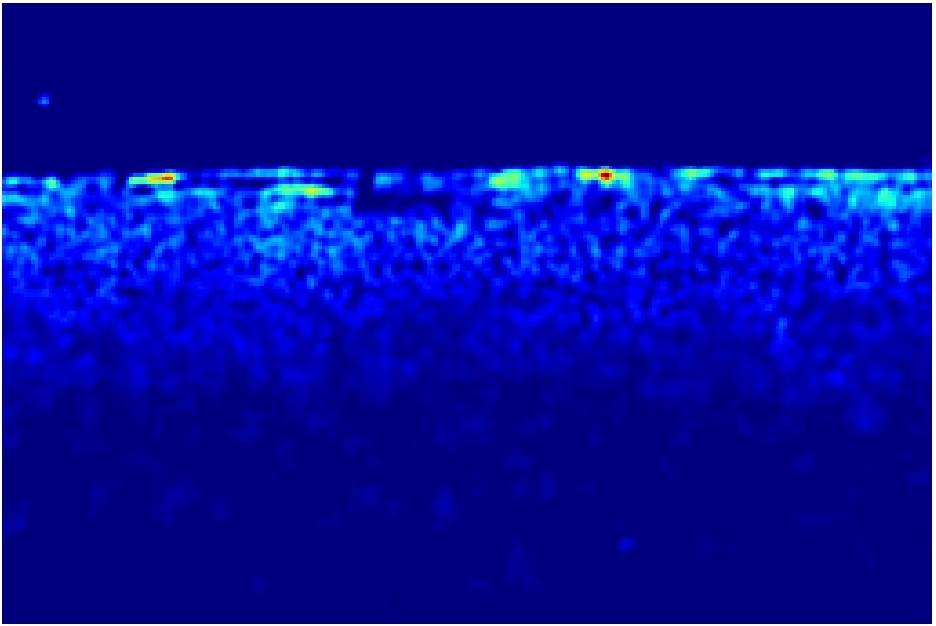} &
  \includegraphics[width=0.27\textwidth]{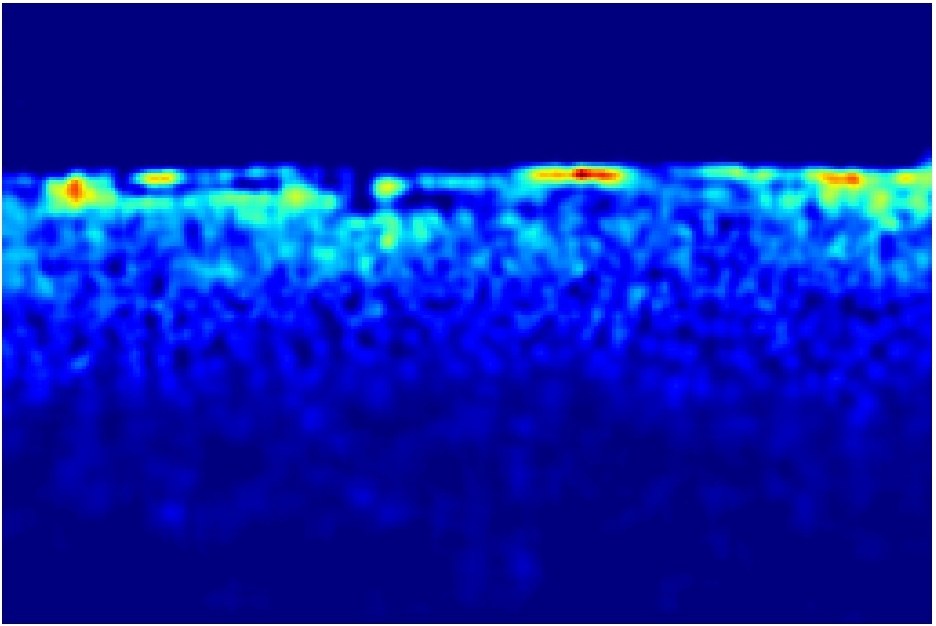} \\
  (d) fused density (1297.1) & (e) attention (S1) & (f) attention (S2) \\
  \includegraphics[width=0.27\textwidth]{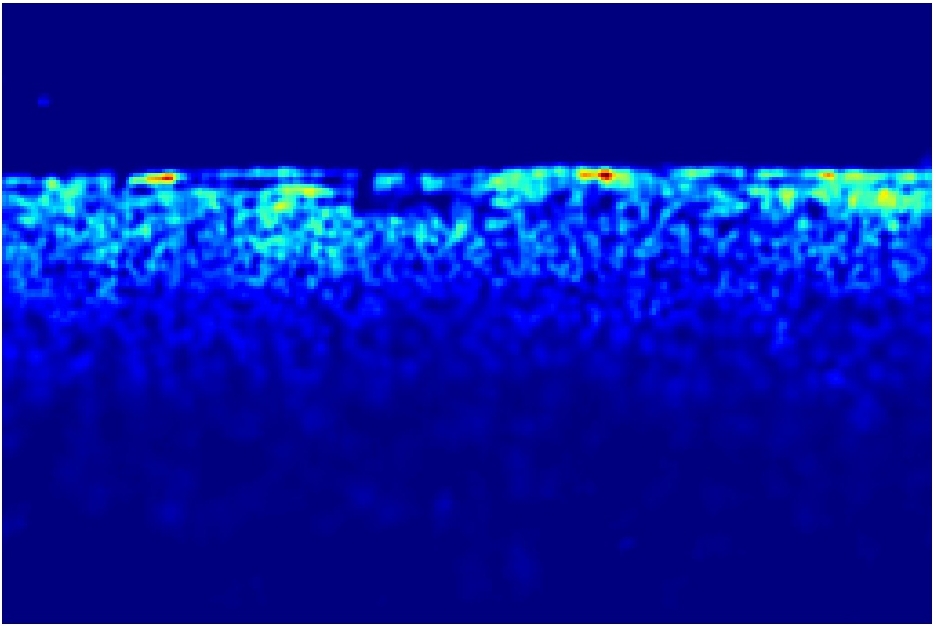} &
  \includegraphics[width=0.29\textwidth]{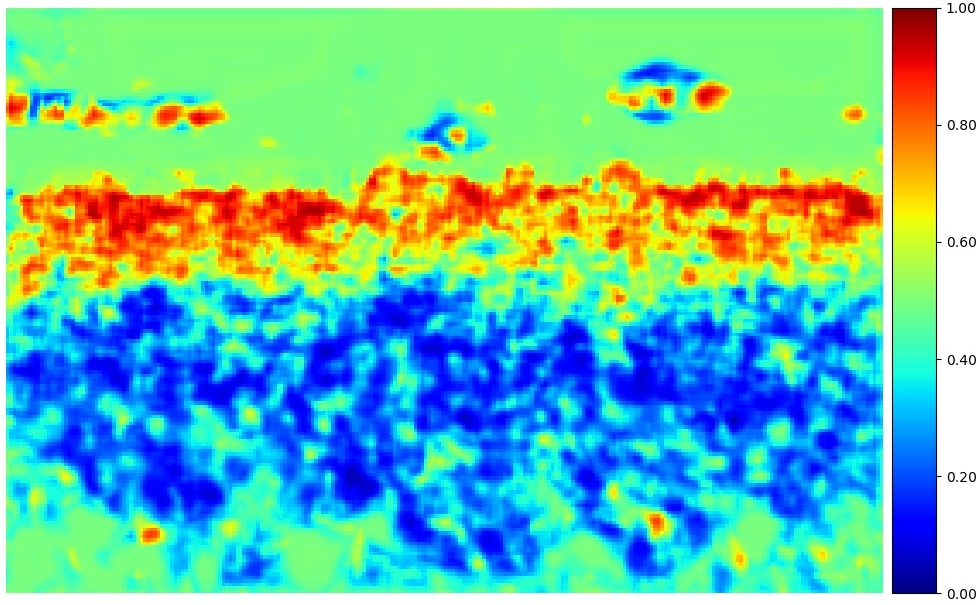} &
  \includegraphics[width=0.29\textwidth]{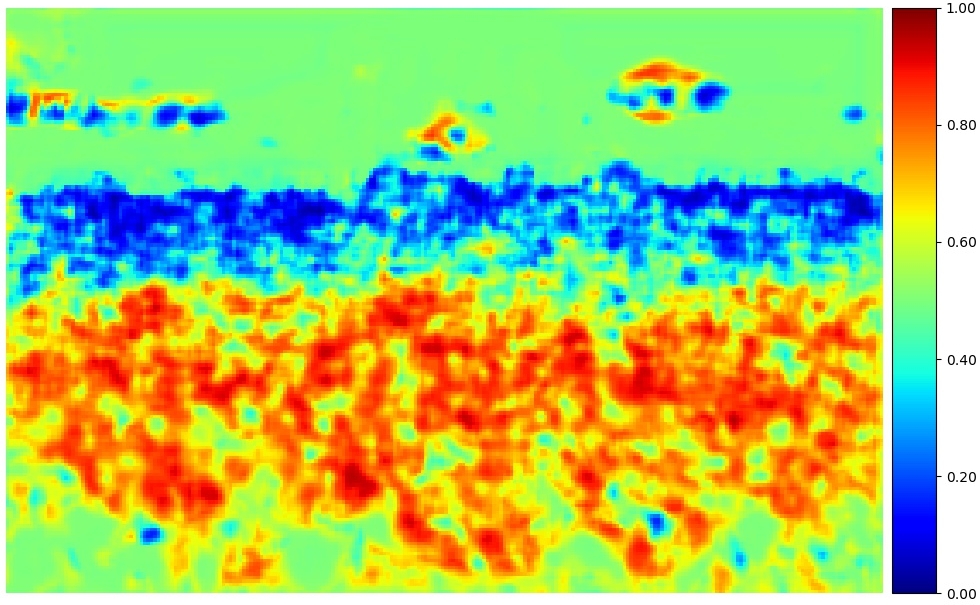} \\
\end{tabular}
\vspace{0.1cm}
\caption{Example of our image pyramid CNN (FCN-7c-2s) showing the predicted attention maps (e, f), density predictions (1-channel feature map) of each scale (b, c) and the final fused density map (d). Numbers inside the parenthesis are the GT density and predicted density. Best viewed in color.
}
\label{fig:demo_ShanghaiTech_A}
\vspace{-0.6cm}
\end{figure}

\par
Most of the recent counting methods adopt object density maps and deep neural networks, due to the counting accuracy and preservation of spatial information in the density map, and the powerful learning capability of deep neural networks.
\cite{Zhang2015Cross,Kang2017Beyond,Walach2016Boost,Onoro2016Hydra} use Alexnet-like~\cite{AlexNet} networks including convolution layers and fully-connected layers. CNN-pixel~\cite{Kang2017Beyond} only predicts the density value of the patch center at one time, resulting in a full-resolution density map, but at the cost of slow prediction.
\cite{Zhang2015Cross,Walach2016Boost,Onoro2016Hydra} predict a density patch reshaped from the final fully-connected layer, resulting in much faster prediction. However, in order to avoid huge fully-connected layers, only density maps with reduced resolution are produced, and artifacts exist at the patch boundaries.
Starting from \cite{Long2015FCN}, fully convolutional network (FCN) becomes very popular for dense prediction tasks such as semantic segmentation~\cite{Long2015FCN}, edge detection~\cite{Xie2015}, saliency detection~\cite{Hou2017} and crowd counting~\cite{Zhang2016MCNN} because FCN is able to reuse shared computations, resulting in much faster prediction speed compared to sliding window-based prediction.

\par
Scale variation (from image to image) coupled with perspective distortion (within one image) is one major difficulty faced by counting tasks, especially for cross-scene scenarios.
To overcome this issue, \cite{Zhang2015Cross} resizes all the objects to a canonical scale, by cropping image patches whose sizes vary according to perspective value, and then resizing them to a fixed size as the input into their network.
However, there is no way to handle intra-patch scale variation for this patch normalization method, and the perspective map needs to be known.
Hydra-CNN~\cite{Onoro2016Hydra} takes the image patch and several of its smaller center crops,
and then upsamples them to the same size before inputing into each of the CNN columns -- the effect is to show several zoomed in crops of the given image patch.
Since the input image contents are different, feature maps extracted from different inputs cannot be spatially aligned. Hence, Hydra-CNN uses a fully-connected layer to fuse information across scales.
In contrast to Hydra-CNN, our proposed method is based on an image pyramid, where each scale contains the whole image at lower resolutions. Hence, the feature maps predicted for each scale can be aligned after upsampling, and a simple fusion using 1x1 convolution can be used.

\par
\cite{Kang2017ACNN} uses side information, such as perspective, camera angle and height, as input to adaptively generate convolution weights to adapt feature extraction to the specific scene/perspective.
However, side information is not always available.
MCNN~\cite{Zhang2016MCNN} uses 3 columns of convolution layers with the same depth but with different filter sizes.
Different columns should respond to objects at different scales, although no explicit supervision is used.
MCNN is adopted by several later works~\cite{Sam2017Switch,Sindagi2017CPCNN}. Switch-CNN~\cite{Sam2017Switch} relies on a classification CNN to classify the image patches to one of the columns to make sure every column is adapted to a particular scale. CP-CNN~\cite{Sindagi2017CPCNN} includes MCNN as a part of their network and further concatenates its feature maps with local and global context features from classification networks.

\par
Despite the success of MCNN, we notice three major issues of MCNN.
Firstly, in order to get a larger receptive field (with fixed depth), $9\times9$ filters are used, which is not as effective as smaller filters \cite{Simonyan14VGG,Szegedy2015Inception,Szegedy2016}. We also experimentally show that a network with similar receptive field size but using larger filters (e.g. $7\times7$ vs $5\times5$) gives worse performance (see Table~\ref{tab:shanghaitech_ablation}).
Secondly, since images/image patches containing objects of all scales are used to train the network, it is highly likely that every column still responds to all scales, which turns MCNN into an ensemble of several weak regressors (but extracting multi-scale features).
To alleviate this problem, \cite{Sam2017Switch} introduces a classification network to assign image patches to one of the three columns, so that only one column is activated every time and trained for a particular scale only. However, the classification accuracy is limited and the true label of an image patch is unclear.
Another issue is that the classification CNN makes a hard decision, resulting in Switch-CNN extracting features from only one scale, thus ignoring other scales that might provide useful context information.
Thirdly, MCNN uses $1\times1$ convolution layer to fuse features from different columns, which is essentially a weighted sum of all the input maps.
However, the weights are the same for all spatial locations, which means the scale with the largest weight will always dominate the prediction, regardless of the image input.
However, to handle scale variations within an image requires each pixel to have its own set of weights to select the most appropriate scale.
This intra-image variation is partially handled by Switch-CNN~\cite{Sam2017Switch} by dividing the whole image into several smaller image patches, which then are processed by a certain column according to the classification result.

\vspace{-0.4cm}
\section{Methodology} \label{sec:model}
\vspace{-0.2cm}

To overcome the issues mentioned above, our solution is to adaptively fuse features from an image pyramid.
Instead of increasing the filter size to see more image content, we send the downsampled images into an FCN with a fixed filter size.
To some extent, this is equivalent to using larger filter size that can cover larger receptive field size, but with less computation and less issues associated with large filters.
Unlike ~\cite{Onoro2016Hydra}, our image pyramid contains the same image content but at different resolutions. Hence, feature maps from different columns can be spatially aligned by upsampling, which is required for  fusion by convolution layers.

\begin{figure}[tb]
\centering
\includegraphics[width=0.98\linewidth]{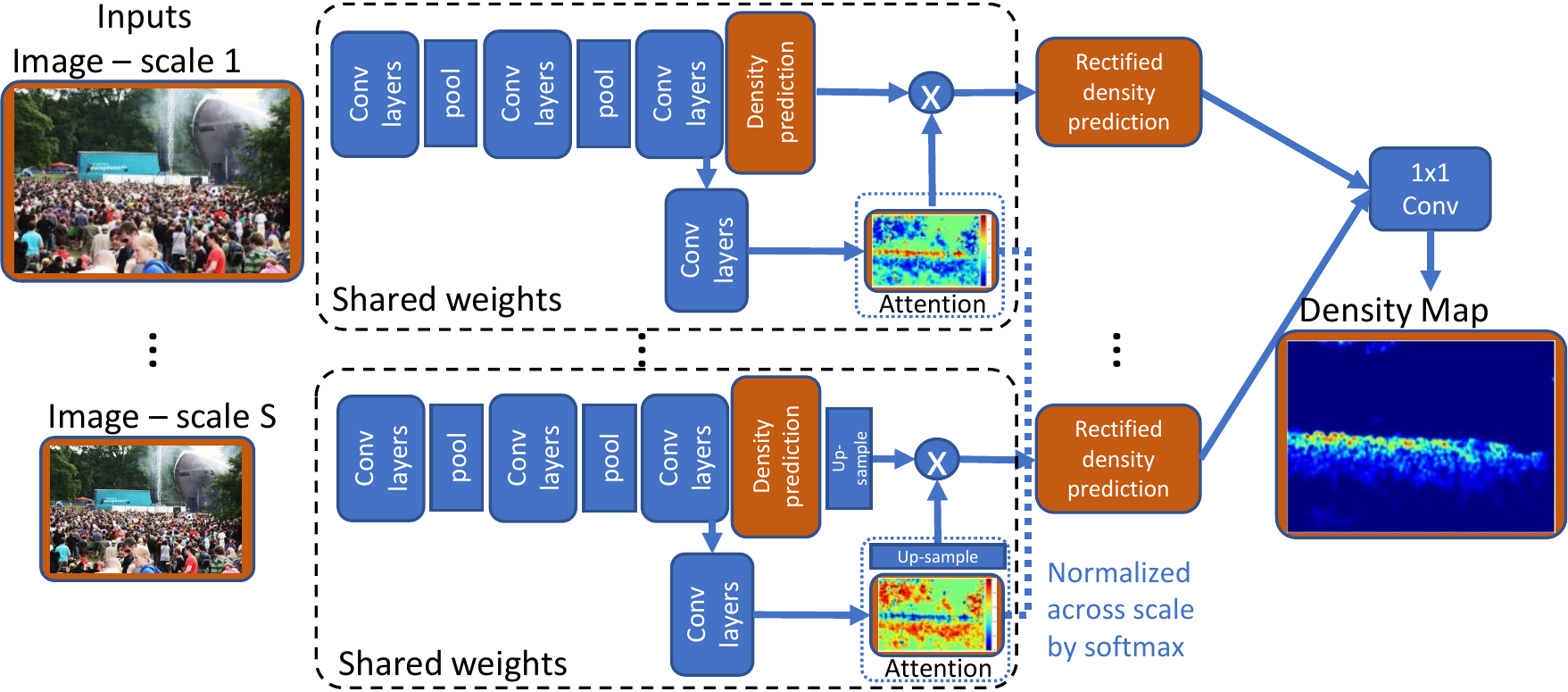}
\vspace{0.2cm}
\caption{
Our image pyramid CNN for crowd counting.
Each scale of the image pyramid is sent into an FCN to predict a density map for that scale. At each scale, the attention sub-net takes as input the last feature map and predicts an attention map. An across-scale softmax is applied to all the attention maps, which is then multiplied to the density maps at each scale. Finally, a 1x1 convolution fuses the density maps from all scales.
The FCN and attention sub-net weights are shared across all scales.
}
\label{fig:arch}
\vspace{-0.6cm}
\end{figure}

Due to scene perspective, the scales of the objects changes based on the location in the image.
Hence, different fusing weights are required for every single location of the feature map. To this end, we estimate an attention map for each scale, which is used to adaptively fuse the predictions.
A similar method is also used in \cite{Chen2016Attention} for semantic segmentation. Experimentally, we find applying attention map on density predictions from different scales is better than applying on feature maps (see Table~\ref{tab:shanghaitech_ablation}).
Our proposed network is shown in \reffig{fig:arch}.

\vspace{-0.3cm}
\subsection{Backbone FCN}
\vspace{-0.2cm}

Given an input image, an image pyramid is generated by downsampling the image to other scales.
Each scale is then input into a backbone FCN to predict a density map for that scale. The backbone FCN uses shared parameters for all scales.

\par
Our backbone FCN contains two max-pooling layers, which divides the whole FCN into three stages and results in predicted density maps downsampled by 4.
The number of convolution layers depends on the general object size of the dataset (7 for ShanghaiTech and 5 for UCSD). Only two pooling layers are used because the resolution of the density map is very important, and it is not easy to compensate for the lost spatial information in overly downsampled density maps~\cite{Kang2017Beyond}.
With only two pooling layers (limited stride size), we use $5\times5$ filters in order to get large enough receptive field size. For example, the FCN-7c used for ShanghaiTech and WorldExpo dataset has a receptive field size of 76.
Table~\ref{tab:FCN_layers} lists the layer configurations.
The final prediction layer uses ReLU activation, while the other convolution layers use leaky ReLU~\cite{Maas2013Leaky} activation (slope of 0.1 for the negative axis).

\begin{table}[tb]

\begin{minipage}[m]{0.6\linewidth}
\centering
\small
\begin{tabular}{|c|c|c|c|}
  \hline
  \multicolumn{2}{|c|}{FCN-7c} & \multicolumn{2}{|c|}{FCN-5c} \\
  \hline
  Layer 	& Filter   		   & Layer 	& Filter  \\
  \hline
  1         & 16x1x5x5    		& 1         & 16x1x5x5  	\\
  2         & 16x16x5x5   		& -			& -  			\\
  max-pooling & 2x2				& max-pooling & 2x2  		\\
  \hline
  3         & 32x16x5x5   		& 2 		& 32x16x5x5  	\\
  4         & 32x32x5x5   		& -         & -			  	\\
  max-pooling & 2x2				& max-pooling & 2x2  		\\
  \hline
  5         & 64x32x5x5			& 3         & 64x32x3x3  	\\
  6         & 32x64x5x5			& 4         & 32x64x3x3  	\\
  7         & 1x32x5x5			& 5         & 1x32x3x3   	\\
  \hline
\end{tabular}
\vspace{0.2cm}
\caption{Our FCN-7c and FCN-5c networks. The four numbers in the Filter columns represent output channels, input channels and filter size (H, W). Bias term is not shown. }
\label{tab:FCN_layers}
\end{minipage}
\hspace{0.7cm}
\begin{minipage}[m]{0.3\linewidth}
\centering
\small
\begin{tabular}{|p{2cm}|c|}
  \hline
  Layer 	& Filter   \\
  \hline
  1         & 8x32x3x3    \\
  \hline
  2         & 1x8x1x1   \\
  \hline
  upsampling & - \\
  \hline
  across-scale softmax & - \\
  \hline
  \hline
  3 (fusion) & 1xSx1x1
  \\
  \hline
\end{tabular}
\vspace{0.2cm}
\caption{Configuration of our attention subnet, and subsequent layer to fuse predictions from all scales into one. $S$ is the number of scales used.}
\label{tab:attention_layers}
\end{minipage}

\vspace{-0.6cm}
\end{table}

\vspace{-0.3cm}
\subsection{Attention map network and fusion}
\vspace{-0.2cm}

\par
Similar to~\cite{Chen2016Attention}, a network containing only two convolution layers is used to generate attention weights from its precedent feature maps, summarizing the information into a 1-channel attention map (see Table~\ref{tab:attention_layers} for details).
The feature maps from the last convolution layer (e.g. conv-6 output for FCN-7c) are used to generate the attention map because they are more semantically meaningful than the low-level features, which respond to image cues.

\par
To perform fusion, the attention maps and the density maps are bilinearly upsampled to the original resolution.
Next, an across-scale softmax normalization is applied on the attention maps so that, at every location, the sum of the attention weights is 1 across scales.
At each scale, the corresponding normalized attention map is element-wise multiplied with the density map prediction for that scale, resulting in a rectified density map where off-scale regions are down-weighted.
Finally, a $1\times1$ convolution layer is applied to fuse the rectified density maps from all scales into a final prediction.
Note that the density predictions (1-channel feature map) from each scale may not strictly correspond to the ground-truth density map, since the fusion step uses a $1\times1$ convolution with bias term and its weights are not normalized.

\vspace{-0.3cm}
\subsection{Training details}
\vspace{-0.2cm}

The whole network is trained end-to-end. The training inputs are 128$\times$128 grayscale randomly cropped image patches (for the original scale), and outputs are 32$\times$32 density patches (every pixel value is the sum of a 4$\times$4 region on the original density patch).
The loss function is the per-pixel MSE between the predicted and ground-truth density patches.
Since the density value is a small fractional number (e.g. 0.001), we multiply it by 100 during training.
SGD with momentum (0.9) is used for training. Occasionally, when SGD fails to converge,
we use Adam~\cite{adam} optimizer for the first few epochs and then switch to SGD with momentum. Learning rate (initially 0.001) decay and weight decay are used.
During prediction, the whole image is input into the network to predict a density map downsampled by a factor of 4.

\vspace{-0.4cm}
\section{Experiment} \label{sec:exp}
\vspace{-0.2cm}

We test our method on three crowd counting datasets: ShanghaiTech, WorldExpo, and UCSD.

\vspace{-0.4cm}
\subsection{Evaluation metric}
\vspace{-0.2cm}

Following the convention of previous works \cite{Chan2008Privacy,Zhang2015Cross}, we compare different methods using mean absolute error (MAE), and mean squared error (MSE) or root MSE (RMSE),
\begin{align}
  \mathrm{MAE}=\tfrac{1}{N}\sum\nolimits_{i=1}^N|\hat{c}_i-c_i|,\
  \mathrm{MSE}=\tfrac{1}{N}\sum\nolimits_{i=1}^N(\hat{c}_i-c_i)^2,\
  \mathrm{RMSE}=\sqrt{\tfrac{1}{N}\sum\nolimits_{i=1}^N(\hat{c}_i-c_i)^2},
\end{align}
where $c_i$ is the ground truth count, which is either an integer number (for UCSD) or fractional number (for other datasets), and $\hat{c}_i$ is the predicted total density which is a fractional number.

\vspace{-0.3cm}
\subsection{ShanghaiTech dataset}
\vspace{-0.2cm}

\begin{table}[tb]

\begin{minipage}[m]{0.49\linewidth}
\centering
\small
\resizebox{\linewidth}{!}{
\begin{tabular}{|@{\hspace{0.1cm}}l@{\hspace{0.1cm}}|c@{\hspace{0.2cm}}c@{\hspace{0.1cm}}|c@{\hspace{0.2cm}}c@{\hspace{0.1cm}}|}
  \cline{2-5}
   \multicolumn{1}{c|}{} & \multicolumn{2}{c|}{Part A}  & \multicolumn{2}{c|}{Part B}  \\
  \hline
  Method                                           & MAE       & RMSE            & MAE    & RMSE  \\
  \hline
  CNN-patch \cite{Zhang2015Cross}                  & 181.8     & 277.7           & 32.0   & 49.8  \\
  MCNN \cite{Zhang2016MCNN}                        & 110.2     & 173.2           & 26.4   & 41.3  \\
  FCN-7c (ours)                                    & 82.3      & 124.7           & 12.4   & 20.5  \\
  FCN-7c-2s (ours)                                 & 81.3      & 132.6           & 10.9   & 19.1  \\
  FCN-7c-3s (ours)                                 & \bf{80.6}      & 126.7           & \bf{10.2}   & 18.3  \\
  \hline
  Switch-CNN \cite{Sam2017Switch}                  & 90.4      & 135.0           & 21.6   & 33.4  \\
  CP-CNN \cite{Sindagi2017CPCNN}                   & \bf{73.6}      & 106.4           & 20.1   & 30.1  \\
  \hline
\end{tabular}
}
\vspace{0.2cm}
\caption{Test errors on the ShanghaiTech dataset. ``2s'' and ``3s'' indicate using 2 or 3 different scales in the image pyramid. Switch-CNN and CP-CNN both use pre-trained VGG~\cite{Simonyan14VGG} network.}
\label{tab:shanghaitech}
\end{minipage}
\hspace{0.15cm}
\begin{minipage}[m]{0.49\linewidth}
\small
\centering
\resizebox{\linewidth}{!}{
\begin{tabular}{|@{\hspace{0.1cm}}l@{\hspace{0.1cm}}|@{\hspace{0.1cm}}c@{\hspace{0.15cm}}c@{\hspace{0.15cm}}c@{\hspace{0.15cm}}c@{\hspace{0.15cm}}c@{\hspace{0.1cm}}|@{\hspace{0.1cm}}c@{\hspace{0.1cm}}|}
  \hline
  Method                            & S1  & S2  & S3  & S4 & S5 & Avg. \\
  \hline
  CNN-patch\cite{Zhang2015Cross} 	& 9.8 & 14.1  & 14.3 & 22.2 & 3.7     & 12.9  \\
  MCNN \cite{Zhang2016MCNN}         & 3.4  & 20.6  & 12.9  & 13.0  & 8.1				& 11.6  \\
  FCN-7c (ours)                     & 2.4 & 15.6 & 12.9 & 28.5 & 5.8					& 13.0  \\
  FCN-7c-2s (ours)                  & 2.0 & 13.4 & 12.1 & 26.5 & 3.3					& 11.5  \\
  FCN-7c-3s (ours)                  & 2.5 &	16.5 & 12.2	& 20.5 & 2.9					& 10.9  \\
  \hline
  Switch CNN~\cite{Sam2017Switch}   & 4.2  & 14.9  & 14.2  & 18.7  & 4.3				& 11.2  \\
  CP-CNN~\cite{Sindagi2017CPCNN}	& 2.9  & 14.7  & 10.5  & 10.4  & 5.8				& 8.9   \\
  \hline
\end{tabular}
}
\vspace{0.2cm}
\caption{MAE on WorldExpo dataset. Only results using the same ground truth densities as the original paper \cite{Zhang2015Cross} are included. Other methods are not comparable since they use different ground-truth counts due to different ground-truth densities.}
\label{tab:worldexpo}
\end{minipage}

\vspace{-0.6cm}
\end{table}

The ShanghaiTech dataset contains two parts, A and B, tested separately.
Part A contains 482 images randomly crawled online, with different resolutions, among which 300 images are used for training and the rest are for testing.
Part B contains 716 images taken from busy streets of the metropolitan areas in Shanghai, with a fixed resolution of $1024\times768$.
400 images are used for training and 316 are for testing. Following \cite{Zhang2016MCNN}, we use geometry-adaptive kernels (with $k$=5) to generate the ground-truth density maps for Part A, and isotropic Gaussian with fixed standard deviation (std=15) to generate the ground-truth density maps for Part B.
We test our baseline FCN, denoted as FCN-7c, as well as 2 versions using image pyramids, FCN-7c-2s uses 2 scales $\{1.0, 0.7\}$ and FCN-7c-3s uses 3 scales $\{1.0, 0.7, 0.5\}$.\footnote{Scale $0.7$ indicates the image width and height are only $70\%$ of the original size.}

\par
Results are listed in Table~\ref{tab:shanghaitech}. Our FCN-7c is a strong baseline, and adaptively fusing predictions from an image pyramid (FCN-7c-2s and FCN-7c-3s) further improves the performance.
On Part B, our method achieves much better result than other methods.
On Part A, our method achieves second best performance, with only CP-CNN performing better.
Note that CP-CNN uses a pre-trained VGG network and performs per-pixel prediction with a sliding window, which is very slow for high-resolution images, running at 0.11 fps (9.2 seconds per image). In contrast, FCN-7c and FCN-7c-3s run at 439 fps and 158 fps (see Table.\ref{tab:shanghaitech_ablation}).

\par
\reffig{fig:demo_ShanghaiTech_A} shows an example results and attention maps from our FCN-7c-2s.
Since scale 1 (S1) takes as input the original image, its features are suitable (higher weight on attention map) for smaller objects. Scale 2 (S2) takes as input the lower resolution image, and responds to the originally larger objects that are now resized down.
More demo images and predictions can be found in the supplemental.

\vspace{-0.3cm}
\subsection{WorldExpo dataset}
\vspace{-0.2cm}

The WorldExpo'10 dataset \cite{Zhang2015Cross} is a large scale crowd counting dataset captured from the Shanghai 2010 WorldExpo.
It contains 3,980 annotated images from 108 different scenes.
Following \cite{Zhang2015Cross}, we use a human-like density kernel, which changes with perspective, to generate the ground-truth density map.
Models are trained on the 103 training scenes and tested on 5 novel test scenes. The network setting is identical to those used for ShanghaiTech.

Results are listed in Table~\ref{tab:worldexpo}.
Adaptive fusion of scales (FCN-7c-2s and FCN-7c-3s) improves the performance over the single scale FCN-7c, and achieves slightly better performance than MCNN and Switch-CNN.
Switch-CNN uses pre-trained VGG weights to fine-tune their classification CNN.
CP-CNN, taking global and local context information into consideration, performs the best.
CP-CNN also uses pre-trained VGG weights when extracting global context features.
Demo images and predictions can be found in the supplemental.

\vspace{-0.3cm}
\subsection{UCSD dataset}
\vspace{-0.2cm}

The UCSD dataset is a low resolution (238$\times$158) surveillance video of 2000 frames, with perspective change and heavy occlusion between objects.
We test the performance using the traditional setting \cite{Chan2008Privacy}, where all frames from 601-1400 are used for training and the remaining 1200 frames for testing. The ground-truth density maps use the Gaussian kernel with $\sigma=4$.
Since the largest person in UCSD is only about 30 pixels tall, our FCN backbone network uses five convolution layers, denoted as FCN-5c (see Table~\ref{tab:FCN_layers}).

Results are listed in Table~\ref{tab:ucsd}. Again, our adaptive fusion improves the performance over FCN-5c, even on this single scene dataset without severe scale variation. Switch-CNN performs worse than either our method or MCNN on UCSD. One possible reason is that the object size is too small compared to ImageNet, and the limited training samples cannot fine-tune their classification CNN well.
See supplemental for our predicted density maps.

\begin{table}[tb]
\begin{minipage}[m]{0.49\linewidth}
\centering
\small
\begin{tabular}{|c|cc|}
  \hline
  Method                                           & MAE   & MSE  \\
  \hline
  CNN-patch \cite{Zhang2015Cross}                  & 1.60  & 3.31  \\
  MCNN \cite{Zhang2016MCNN}                        & \bf{1.07}  & 1.82$^*$ \\
  CNN-boost \cite{Walach2016Boost}   & 1.10  & - \\
  CNN-pixel \cite{Kang2017Beyond}                  & 1.12  & 2.06  \\
  \hline
  FCN-5c (ours)                                    & 1.28  & 2.65  \\
  FCN-5c-2s (ours)                                 & 1.22  & 2.33  \\
  FCN-5c-3s (ours)                                 & 1.16  & 2.29  \\
  \hline
  Switch-CNN \cite{Sam2017Switch}                  & 1.62  & 4.41$^*$  \\
  \hline
  Hydra 3s  \cite{Onoro2016Hydra} (``max'')	   & 2.17  & - \\
  ACNN \cite{Kang2017ACNN}  (``max'')			   & 0.96  & - \\
  \hline
\end{tabular}
\vspace{0.2cm}
\caption{Test errors on the UCSD dataset when using the whole training set. ``max'' means the methods are trained on the downsampled training set, where only one out of 5 frames is used for training.
{\scriptsize $^*$ \cite{Zhang2016MCNN} and  \cite{Sam2017Switch} report RMSE of 1.35 and 2.10.}
}
\label{tab:ucsd}
\end{minipage}
\hspace{0.1cm}
\begin{minipage}[m]{0.49\linewidth}
\centering
\small
\begin{tabular}{|c|c@{\hspace{0.15cm}}c@{\hspace{0.15cm}}c@{\hspace{0.15cm}}c@{\hspace{0.1cm}}|@{\hspace{0.1cm}}c@{\hspace{0.15cm}}|c@{\hspace{0.15cm}}|}
  \cline{2-6}
  \multicolumn{1}{c|}{} & \multicolumn{5}{c@{\hspace{0.15cm}}|}{Rank} & \multicolumn{1}{c}{} \\
  \hline
  Method & \rotatebox{90}{Shanghai Tech A} & \rotatebox{90}{Shanghai Tech B} & \rotatebox{90}{WorldExpo} & \rotatebox{90}{UCSD} & \rotatebox{90}{Average} &  \rotatebox{90}{Runtime (fps)} \\
  \hline
  MCNN~\cite{Zhang2016MCNN} & 4 & 4 & 4 & 1 & 3.25 & 307  \\
  Switch-CNN~\cite{Sam2017Switch} & 3 & 3 & 3 & 3 & 3.00 & 12.9  \\
  CP-CNN~\cite{Sindagi2017CPCNN} & 1 & 2 & 1 & - & 1.33 & 0.11  \\
  Ours (FCN-x-3s) & 2 & 1 & 2 & 2 & 1.75 & 158  \\
  \hline
\end{tabular}
\vspace{0.2cm}
\caption{Comparison of the four most competitive methods.
All runtimes are using PyTorch 0.3.0 on a GeForce GTX 1080 on 1024$\times$768 images. Methods without published code were built (without training) and speed tested in the prediction stage.}
\label{tab:rank}
\end{minipage}
\vspace{-0.5cm}
\end{table}

\vspace{-0.3cm}
\subsection{Summary}
\vspace{-0.2cm}

Overall, MCNN~\cite{Zhang2016MCNN}, Switch-CNN~\cite{Sam2017Switch}, CP-CNN~\cite{Sindagi2017CPCNN} and our proposed method using three scales (FCN-x-3s) perform the best on the tested datasets. We summarize their performance rank and runtime speed in Table~\ref{tab:rank}.
Overall, our method has better average rank than MCNN~\cite{Zhang2016MCNN} and Switch-CNN~\cite{Sam2017Switch}, but is worse than CP-CNN.
Switch-CNN~\cite{Sam2017Switch} and CP-CNN~\cite{Sindagi2017CPCNN} both use pre-trained VGG model and do not run in a fully convolutional fashion, resulting in slower prediction.
CP-CNN is especially slow because it uses sliding-window per-pixel prediction to get the local context information.
In contrast, MCNN~\cite{Zhang2016MCNN} and our FCN-x-3s are fully convolutional, resulting in faster than real-time prediction.
They also use much smaller models and do not need extra data to pre-train the models (c.f., VGG).

\vspace{-0.3cm}
\subsection{Ablation study and discussion}
\vspace{-0.2cm}

In this section, we discuss the design of our model with ablation studies.
All the ablation study results tested on ShanghaiTech are summarized in Table~\ref{tab:shanghaitech_ablation}, and Table~\ref{tab:ucsd_ablation} for UCSD.

\vspace{-0.3cm}
\subsubsection{Backbone FCN design} \label{sec:FCN_arch_choice}
\vspace{-0.2cm}

\begin{table}[tb]
\centering
\footnotesize
\begin{tabular}{|c|c|c|c|c|c|c|c|}
  \cline{4-7}
        \multicolumn{3}{c|}{}    				& \multicolumn{2}{c|}{Part A}  & \multicolumn{2}{c|}{Part B} & \multicolumn{1}{c}{} \\
  \hline
  Method                  & RF & \# parameters  & MAE       & RMSE            & MAE    & RMSE  	& runtime (fps) \\
  \hline
  FCN-7c (ours)           & 76 & 148,593       & 82.3      & 124.7           & 12.4   & 20.5  	& 439                  \\
  \hline
  FCN-7c-40               & 40 & 162,337       & 106.0     & 153.4           & 19.8   & 33.5  	& 433                 \\
  FCN-7c-40 (scale $0.7$) & 40 & 162,337       & 97.2      & 155.9           & 17.1   & 23.9  	& -                 \\
  \hline
  FCN-5c-78               & 78 & 178,369       & 101.8     & 137.9           & 15.5   & 26.8  	& 494                  \\
  FCN-14c-76              & 76 & 178,369       & 83.4      & 142.4           & 12.7   & 19.0  	& 329                  \\
  \hline
  FCN-7c-3s (ours)        & - & 150,918        & 80.6      & 126.7           & 10.2   & 18.3  	& 158                 \\
  FCN-7c-3s (fixed)       & - & 148,597   & 85.8      & 135.8           & 10.3   & 16.7 	& -					\\
  FCN-7c-3s (w/o softmax) & - & 150,918   & 87.5 	  & 130.0 			& 11.0   & 16.6     & -     \\
  FCN-7c-3s (sum)		  & - & 150,914   & 83.8 	  & 120.1 			& 10.4   & 16.8     & -     \\
  FCN-7c-3s (low)         & - & 149,182   & 85.6      & 133.8           & 11.0   & 19.2   	& -					\\
  FCN-7c-3s (feat)        & - & 150,210   & 82.0      & 126.6           & 11.1   & 19.0   	& -					\\
  \hline
  FCN-7c-2s (ours)        & - & 150,917        & 81.3      & 132.6           & 10.9   & 19.1  	& 234                 \\
  FCN-7c-2s (fixed)       & - & 148,596        & 86.6      & 130.8           & * & *	& -					\\
  FCN-7c-2s (w/o softmax) & - & 150,917		   & 84.3	   & 132.2			 & 11.0   & 18.5    & -     \\
  FCN-7c-2s (sum)		  & - & 150,914		   & 82.8	   & 123.4			 & 11.3   & 19.4    & -     \\
  FCN-7c-2s (low)         & - & 149,181        & 82.8      & 129.4           & 12.2   & 19.0  	& -					\\
  FCN-7c-2s (feat)        & - & 150,178        & 90.0      & 137.5           & 12.4   & 22.0  	& -					\\
  \hline
\end{tabular}
\vspace{0.2cm}
\caption{Comparison of different variants on ShanghaiTech dataset. RF is the receptive field size.
* indicates that the network failed to converge in limited trials.}
\label{tab:shanghaitech_ablation}
\vspace{-0.5cm}
\end{table}

We find that receptive field (RF) plays a very crucial role in the density estimation task.
The RF needs to be large enough so that enough portion of the object is visible to recognize it. On the other hand, if the RF is too large, then too much irrelevant context information will distract the network.
Hence, the most suitable RF size should be related to the average object size of the dataset.
We conduct several experiments to show the importance of the RF.

\mysubsubsection{Receptive field size}
On ShanghaiTech dataset, we test another FCN with 7 convolution layers whose filter sizes are $3\times3$ for all the convolution layers, resulting in RF size only equal to 40 (denoted as FCN-7c-40).
In order to make a fair comparison, we increase the channel numbers to balance the total parameters. Specifically, the 7 convolution layers now use $\{32,32,64,64,96,48,1\}$ filter channels respectively now.
Although the density values are mostly assigned to head and shoulder regions, but considering the image size (e.g. $1024\times768$ for ShanghaiTech B), an RF of only 40 is too small, and only achieves 106.0 MAE on part A and 19.8 MAE on part B.
However, if this FCN-7c-40 model is trained and tested on scale 0.7, its performance increases to 97.2 MAE on part A and 17.1 MAE on part B.

\par
On UCSD, we test two variants: 1) FCN-5c with filter size of $\{5,5,5,5,5\}$, resulting in RF size of 64 (denoted as FCN-5c-64);
2) FCN-7c used for ShanghaiTech/WorldExpo whose RF size is 76. These two models have more capacity (trainable parameters) than the proposed FCN-5c but achieve worse MAE (1.40 for FCN-5c-64 and 1.54 for FCN-7c), since an RF size of 64 or 76 is too large on UCSD where the largest object is only about 30 pixels tall.

\mysubsubsection{Filter size}
Smaller filter size is preferred since it saves parameters and computation~\cite{Simonyan14VGG,Szegedy2015Inception,Szegedy2016}. Here we experimentally show that the $7\times7$ filter used in MCNN is not the best choice. We test a FCN with 5 convolution layers, whose filter sizes are $\{7,7,7,5,5\}$ respectively, resulting in RF size equal to 78 (denoted as FCN-5c-78) on ShanghaiTech. A substantial performance drop is observed (101.8 MAE on part A and 15.5 MAE on part B).

\par
Since $3\times3$ filters is most commonly used, we replace the $5\times5$ filters in our FCN-7c with 2 convolution layers using $3\times3$ filters. In order to make a fair comparison, we increases the filter number to balance the total trainable parameters. The resulting FCN-14c-76 uses $\{24,24,24,24,32,32,32,32,64,64,32,32,32,1\}$ filters respectively, in total 143,225 parameters (FCN-7c uses 148,593 parameters). It achieves 83.4 MAE on part A and 12.7 on part B (see Table~\ref{tab:shanghaitech_ablation}). Since no performance improvement is observed, we would prefer to use our FCN-7c, considering that deeper network are normally more difficult to train and FCN-14c-76 runs slower than FCN-7c during prediction stage (329 fps vs 439 fps on part B).

\vspace{-0.3cm}
\subsubsection{Fusion design}
\vspace{-0.2cm}

Here we consider different variants of fusing two scales.
We test simple 1$\times$1 convolution without attention-map adaptivity to fuse predictions from different scales (denoted as ``FCN-7c-2s (fixed)''), similar to MCNN~\cite{Zhang2016MCNN}.
On ShanghaiTech, it only achieves 86.6 MAE on part A and failed to converge on part B, showing the necessity of adaptivity during fusion.
We also test 2 other variations of our network: 1) without across-scale softmax normalization (denoted as ``FCN-7c-2s (w/o softmax)''); 2) using simple summation to fuse the density predictions instead of $1\times$1 convolution layer (denoted as ``FCN-7c-2s (sum)''). These variants also give worse MAE than our FCN-7c-2s.

To generate the attention map, we consider using feature maps from other convolution layers, such as the last layer of stage 1. Since its resolution is higher, we insert two average pooling layer after both the convolution layers in the attention sub-network, denoted as ``FCN-7c-2s (low)''. It gives worse performance than FCN-7c.
One possible reason is that lower layer features, although containing more spatial information, are semantically weak, e.g. more similar to edge features. However, the attention map requires high-level abstraction to distinguish the foreground objects at the desired scale from the background.

Other than fusing the density predictions from different scales, similar to MCNN~\cite{Zhang2016MCNN}, we can also concatenate the feature maps and then predict a final density map. In this method denoted as ``FCN-7c-2s (feat)'', the generated attention map from the last convolution layer of stage 3 is applied on its output, similar to spatial transformer~\cite{Jaderberg2015Spatial}. However, it does not perform as well as our density fusion.

\begin{table}[tb]
\centering
\small
\begin{tabular}{|c|c|c|c|c|}
  \hline
  Method           & RF & \# parameters   & MAE   & MSE  \\
  \hline
  FCN-5c (ours)    & 40 &  50,497         & 1.28  & 2.65  \\
  FCN-5c-64        & 64 &  116,545        & 1.40  & 3.16  \\
  FCN-7c           & 76 &  148,593        & 1.54  & 3.80  \\
  \hline
  FCN-5c-2s (ours) & - &  52,821       & 1.22  & 2.33  \\
  FCN-5c-3s (ours) & - &  52,822       & 1.16  & 2.29  \\
  \hline
\end{tabular}
\vspace{0.2cm}
\caption{Comparison of different variants on UCSD dataset.
}
\label{tab:ucsd_ablation}
\vspace{-0.6cm}
\end{table}

\vspace{-0.3cm}
\section{Conclusion}
\vspace{-0.2cm}

By taking into consideration the design of backbone FCN and the fusion of predictions from an image pyramid,
our proposed adaptive fusion image pyramid counting method achieves better average ranking on 4 datasets than MCNN and Switch-CNN.
Although CP-CNN has better ranking than our proposed method, it is much slower since it uses both a very deep network (VGG) and sliding window per-pixel prediction.
Taking into consideration of model size and running time, our method is the most favorable, especially for cases requiring real-time prediction.

\vspace{0.2cm}

\noindent
{\bf Acknowledgement}:
This work was supported by grants from the Research Grants Council of the Hong Kong Special Administrative Region, China (Project No.\,[T32-101/15-R] and CityU 11212518), and by a Strategic Research Grant from City University of Hong Kong (Project No.\,7004887). We grateful for the support of NVIDIA Corporation with the donation of the Tesla K40 GPU used for this research.

\bibliography{bib_BMVC_2018}
\end{document}